# Almost Optimal Intervention Sets for Causal Discovery

Frederick Eberhardt*
Institute of Cognitive and Brain Sciences
University of California, Berkeley
Berkeley, CA 94720; and

## Abstract

We conjecture that the worst case number of experiments necessary and sufficient to discover a causal graph uniquely given its observational Markov equivalence class can be specified as a function of the largest clique in the Markov equivalence class. We provide an algorithm that computes intervention sets that we believe are optimal for the above task. The algorithm builds on insights gained from the worst case analysis in Eberhardt et al. (2005) for sequences of experiments when all possible directed acyclic graphs over $N$ variables are considered. A simulation suggests that our conjecture is correct. We also show that a generalization of our conjecture to other classes of possible graph hypotheses cannot be given easily, and in what sense the algorithm is then no longer optimal.

## 1 INTRODUCTION

Suppose we have a set of variables $\mathbf{V} = \{X_1, \ldots, X_n\}$ and we are interested in discovering the causal relations among these variables. That is, we want to find out which variable influences which other variable directly or indirectly. This problem is common in both the natural sciences (e.g. gene regulatory networks, drug testing etc.) and the social sciences (e.g. econometrics, policy decisions etc). Depending on the area of research the scientist will have different tools to investigate this problem and different model space assumptions and background knowledge will be appropriate or available. The main problem for the scientist is to determine and justify these assumptions and to use tools of investigation that will distinguish the true causal structure from other possible causal structures, i.e. the scientist must reduce and, if possible, rule out underdetermination of the causal structure.

For any particular set of causal variables, the scientist will consider a set of possible hypotheses that describe the causal relations among these variables. Causal Bayes nets (Spirtes et al. 2000; Pearl 2000) provide a concise framework by representing causal structures in terms of directed acyclic graphs (DAGs) connecting the variables, and a probability distribution over these variables, that factors according to the DAG. If nothing is known about the causal structure then the set of hypotheses is large, namely, all possible DAGs over the set of variables, including maybe structures that involve latent variables. For these cases search procedures can and have been devised (see Eberhardt 2007 for details and references) that – given a set of assumptions over the model space – determine the causal structure uniquely, or give a precise specification of the remaining underdetermination. In particular, the following result is relevant to this paper. In Eberhardt et al. (2005) we showed that:

**Theorem 1.1** (Worst Case Bound for Complete Hypothesis Space). *Given a set of $N$ causally sufficient variables $\lfloor \log_2(N) \rfloor + 1$ experiments are sufficient and in the worst case necessary to discover the causal structure uniquely if multiple variables can be subject to an intervention simultaneously and independently.*

That is, if all that is known about the true causal structure among $N$ variables is that there are no latent common causes, then the structure can be determined uniquely in at most $\lfloor \log_2(N) \rfloor + 1$ experiments, where each experiment may involve a randomization of several variables. The bound is based on the combinatorics of different ideal experiments, assuming one has access to an oracle for the independence constraints of the (manipulated) distribution, i.e. it applies to the large sample limit.

But in actual science, it is rarely the case that the

---

*Also: Department of Philosophy, Washington University in St. Louis, St. Louis, MO 63130; Contact: fde@berkeley.edu.

hypothesis set is so general and uninformative. Often substantial knowledge about parts of the causal structure is already available, and only a search among a restricted set of hypotheses is required. In this case the search procedures implied by the worst case bound are overkill. Something more adaptive is required. Take a simple case, where we only have three variables $X, Y$ and $Z$ and suppose we are sure (for whatever reason) that they can only be causally related as in one of the following two DAGs:

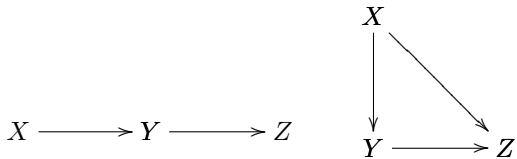

Suppose, for the sake of argument, that no other variables – measured or unmeasured – are relevant, i.e. $X, Y$ and $Z$ are causally sufficient. In this case, the search procedure is simple: The cheapest way of distinguishing the DAGs is probably to collect passive observational data: If $X \perp\!\!\!\perp Z | Y$ then the first structure is true, otherwise the second. But as is well known, passive observation does not distinguish all possible causal structures. In particular, the following three structures cannot be distinguished based alone on the independence relations they imply:

$$X \to Y \to Z \quad X \leftarrow Y \to Z \quad X \leftarrow Y \leftarrow Z$$

Together they form a so-called observational Markov equivalence class (OME) of causal structures, where each DAG in the OME implies the same independence constraints. There are more and less obvious ways to proceed: A less obvious one is that if the causal relations among the variables form a linear structural equation model with *non*-Gaussian error terms, then these three causal hypotheses can be distinguished with passive observational data alone (Shimizu et al. 2006). The more obvious procedure is to perform an experiment: If we intervene with a randomization on $Y$ then we are able to uniquely distinguish the three hypotheses: If the first one is true, then under the randomization of $Y$ we have $X \perp\!\!\!\perp Y$, since the randomization makes the intervened variable ($Y$) independent of its normal causes ($X$ in this case). This feature of randomization is what is sometimes referred to as "edge-breaking". We do not find this independence for the other two structures under the same manipulation. For similar reasons, if the third hypothesis is true, then $Y \perp\!\!\!\perp Z$ under a randomization of $Y$, whereas under the same circumstances $Y \not\!\perp\!\!\!\perp Z$ for hypotheses one and two.

Obviously, if the hypothesis set were different, then an experimental intervention on $Y$ would not necessarily distinguish (all) the hypotheses. In the simplest case, suppose we just have two variables and just two possible hypotheses: Either $X$ causes $Y$ or the two variables are causally separate. In that case an intervention on $Y$ would be useless, since under a randomization of $Y$ the two structures are equivalent: Both imply that $X \perp\!\!\!\perp Y$ given that $Y$ was randomized. Depending on the hypothesis space and the sequence of experiments, underdetermination can be difficult to resolve. So, analogously to an OME, we refer to structures that imply the same independence constraints in a sequence of experiments as forming a manipulated Markov equivalence class (MME) for that sequence of experiments.

Given OMEs and MMEs as common examples of more restricted hypothesis spaces, the obvious question is whether we can provide efficient search procedures that ensure that we reduce the underdetermination as fast as possible. Can we state guarantees on those search procedures, just as Theorem **??** gives guarantees for search procedures on the complete hypothesis space?

## 1.1 BACKGROUND

Traditionally, causal discovery algorithms split into two categories: constraint based and score based algorithms. Score based algorithms, such as the GES-algorithm (Chickering 2002), compute a score for each model given the measured data. The model with the highest score is then deemed to be the most likely or most plausible model given the data. In contrast, constraint based algorithms (e.g. the PC-algorithm (Spirtes et al. 2000)) test for particular constraints in the measured data and select models that match the discovered constraints. Most commonly, independence constraints are used, but many other constraints can also be considered.

Without further assumptions (such as time order, non-Gaussian errors etc.) both types of search algorithms are asymptotically limited to the discovery of OMEs or MMEs of causal structures. But as structure search algorithms they are uninformative about which experiment(s) should be performed, if a sequence of experiments is necessary to further reduce underdetermination. Tong & Koller (2001), Murphy (2001) and Cooper and Yoo (1999) have presented Bayesian approaches on how to select the next experiment *given* an OME or MME. Generally speaking, all of their approaches are based on selecting the next experiment that maximally reduces the entropy among the remaining hypotheses. This is computationally extremely expensive since an optimization over all possible parameterizations of all possible models has to take place. All of the cited approaches use different computational shortcuts to approximate the computation.

A similar procedure based on the qualitative graphical structure, which would be much more suited for constraint based algorithms, has not been given, although various suggestions for measures of selecting the next best experiment in such circumstances have been made in Meganck et al. (2005). We are unaware of any algorithmic implementations of how to compute the intervention sets efficiently given the measures, or of bounds on the search procedure that these measures imply. The advantage of an algorithm for the selection of experiments based only on qualitative features of the causal structure is that it does not require any distribution over the hypothesis space or any parameterization of the causal structures, both of which may only be rarely available at the beginning of actual scientific inquiries.

### 1.2 CLASSES OF HYPOTHESES

There are many ways in which information may be available that restricts the hypothesis set of causal structures. We will use a *knowledge graph* to characterize types of structural constraints that restrict the hypothesis set.

**Definition 1.2** (Knowledge Graph). *A knowledge graph is a mixed graph over a set of vertices* **V** *such that any two vertices are connected by exactly one of the following edge-types:*

**direct cause:** A directed edge represents the knowledge that one vertex, the start, is a direct cause of the other, the end (relative to **V**): $X \longrightarrow Y$

**non-adjacency:** The absence of an edge represents the knowledge that neither vertex is a direct cause of the other: $X \quad Y$

**adjacency:** An undirected edge represents the knowledge that there is a direct causal connection between the two vertices, but that it is not known in which direction it goes: $X \longrightarrow Y$

**semi-directed:** A semi-directed edge from vertex $X$ to $Y$ represents the knowledge that neither variable is a direct cause of the other or that $X$ is a direct cause of $Y$: $X \dashrightarrow Y$

**no knowledge:** A no-knowledge edge represents the lack of any knowledge about the direct connection between the two vertices: $X \mathrel{-?-} Y$

*An edge in a knowledge graph is considered* known *if it is of one of the first two edge-types, otherwise it is unknown. A knowledge graph is said to represent a* causal structure *uniquely when each of its edges is* known *and the structure is acyclic.*

A knowledge graph, like a pattern for an OME, can be used to represent an equivalence class of graphs that imply the same independence constraints. The main difference to a pattern is that a knowledge graph can represent information about independence relations resulting from interventions. Clearly, an OME of a causally sufficient set of variables can be represented as a knowledge graph, since only the first three edge-types are required. The fourth edge-type is used when exactly one variable of a pair is subject to an intervention and the pair is found to be independent in the experimental data set. The fifth edge-type is used when no information is available about the causal relation between two variables, for example when two variables are subject to interventions simultaneously. Knowledge graphs can represent OMEs and MMEs, but not all knowledge graphs represent classes of graphs that are OMEs or MMEs. For example, for three variables, if each pair is connected by a semi-directed edge, then we would have a knowledge graph, but no passive observation or sequence of one or more experiments would ever yield such an equivalence class.

### 1.3 EXAMPLE

We can now illustrate with an example how intervention sets should be selected when the hypothesis space is restricted. Suppose that for a set of four variables $W, X, Y, Z$, the following knowledge graph is known – it could be the output of a causal search algorithm given data generated in an experiment where $X$ was randomized:

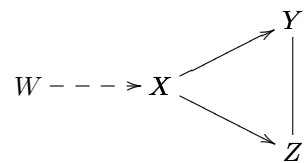

Clearly, the next intervention should be an intervention on $Y$ or $Z$ (and possibly, but not necessarily $W$ also). To resolve the orientation of the undirected edge we must intervene on one of its endpoints: independence between the endpoints will indicate an incoming edge on the intervened endpoint (given the knowledge graph), while dependence for all conditioning sets will indicate an edge outgoing from the intervened endpoint. The semi-directed edge can be resolved if both its endpoints are passively observed or if $W$ is subject to an intervention. In both cases, if $W$ and $X$ are found to be dependent, then – given the knowledge graph – we know that $W \to X$, otherwise we know that there is no direct causal link between the variables.

This was easy. Now suppose we have an OME over the set of variables $\mathbf{V} = \{V, W, X, Y, Z\}$ represented

by the following knowledge graph:[1]

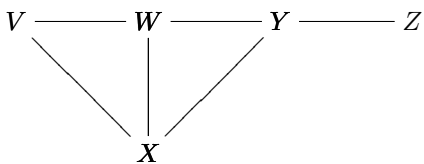

This knowledge graph represents 12 distinct causal hypotheses (substantially less than the 29,281 possible DAGs over 5 variables). Without further assumptions, passive observational data is not going to help. But we can intervene, the question is where? Note that if we can orient one edge, we can determine additional orientations that are implied (see Meek Rules in Meek 1995). For example, if we find that $Z \to Y$, then this also implies $Y \to W \to V$ and $Y \to X \to V$. But intuitively, intervening on $Z$ might seem like a bad idea given this OME. $X$ or $W$ may appear to be better candidates, and rightly so: The advantage of intervening on $X$ is that there are particular structures (two in fact) in the equivalence class, that, if true, would be uniquely distinguished within the equivalence class by a single experiment randomizing $X$. The following is one such structure:

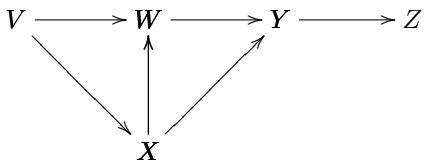

The same is true, albeit for other structures, for interventions on $V, W$ and $Y$, but not for an intervention on $Z$: No matter what the true structure is, an intervention on $Z$ will not uniquely resolve this Markov equivalence class. Further, brief inspection shows that there is no intervention set (of any size) that *guarantees* discovery of the true causal structure in one experiment.

How does one compute intervention sets for knowledge graphs for which intuitions fail us?

The algorithm we propose builds on insights from the analysis underlying Theorem **??**. The key difficulty in uniquely identifying the true causal structure is to determine the orientation of edges in cliques. Cliques are subsets of the vertex set for which every pair of vertices is connected by an edge in the true causal structure. A clique makes edge-orientations maximally independent, because fewer orientations are implied (there are no v-structures[2]; only acyclicity constraints

---
[1]We use the example from Meganck et al. (2005).
[2]Variables $X, Y, Z$ form a v-structure if $X \to Y$ and $Z \to Y$ and there is no direct causal connection between $X$ and $Z$. V-structures are also known as unshielded colliders, and due to their unique independence implications, can be identified in passive observational data.

imply orientations). An algorithm aiming to minimize the number of experiments must therefore break down cliques of connected variables as fast as possible. For a clique of size $|C|$, an experiment that intervenes on $k$ variables in the clique determines the orientation of $k(|C| - k)$ of the clique's edges. This value is maximized for $k = |C|/2$. So our algorithm, presented below, works as follows:

Given the knowledge graph it determines all maximal cliques of variables connected by so-called *unknown* (undirected, semi-directed or no-knowledge) edges. In our example we have: $C_1 = \{V, W, X\}, C_2 = \{W, X, Y\}$ and $C_3 = \{Y, Z\}$. However, we only have to consider $C_1$ and $C_2$, since orientation of all edges in a 3-clique will take 2 experiments in the worst case, and so we can leave resolution of $C_3$ to the second experiment, since it only requires a single experiment (and may anyway get resolved for free in the first experiment). Ideally we should intervene on $|C_i|/2$ variables simultaneously for each clique $C_i$. However, since the cliques may (and in our case do) overlap, the selection of the variables to subject to interventions cannot be done for each clique independently. Consequently, we count for each vertex how many cliques it is part of to determine which vertices are in the most cliques. By considering maximal cliques in order of size and their vertices in order of their prevalence in the cliques, the algorithm greedily approximates the optimal choice of intervention set. In our example, $W$ and $X$ are in 2 (relevant) cliques, while $V$ and $Y$ are only in one. The choice between $X$ and $W$ is random, since there is no a priori reason to prefer one over the other. Once, say, $X$ is selected, all cliques are updated as to whether a sufficient number of their vertices are contained in the intervention set (see algorithm for details). In our case, the selection of $X$ for the intervention set would imply that $C_1$ and $C_2$ have sufficient vertices in the intervention set, since inclusion of any additional vertex from those cliques would not necessarily reduce the number of experiments needed. (Note, that one could include $Z$ in the intervention set to resolve $C_3$, but one need not.) Suppose that upon intervening on $X$ we find the same independence constraints as in the OME. We can then update our knowledge graph to reflect the intervention test:

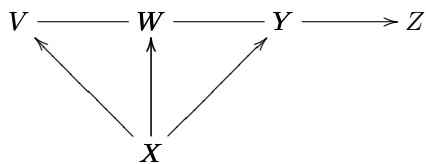

We were unlucky; a further experiment is required, but note that the orientation of the $YZ$-edge was implied by the orientation of the $XY$-edge resulting from the intervention. Now only the $V$–$W$–$Y$ connection has to be resolved: $W$ is added to the new intervention set for the next experiment, since it is the only vertex that is part of two cliques. We know from the discussion earlier that this intervention will be sufficient to uniquely determine the true graph. Our algorithm determined intervention sets that could have resolved the OME in a single experiment, but that ensured that it was resolved in what we knew to be the worst case theoretical bound for this particular knowledge graph: two experiments.

## 2 THEORY

We conjecture that the above procedure works quite generally for knowledge graphs that are OMEs, and that the number of experiments that are in the worst case necessary, is a function of the largest clique in the OME:

**Conjecture 2.1** (Experiments on OMEs). *Given an OME of the true graph, if multiple simultaneous and independent interventions can be performed in each experiment, then $\lceil \log_2(|C_{\max}|) \rceil$ experiments are sufficient and in the worst case necessary to recover the true causal graph, where $C_{\max}$ is the largest clique in the OME.*

Depending on the size of the largest clique in the OME, this bound is substantially lower than the bound of Theorem **??**. But if the largest clique in the OME contains all $N$ variables, then the bounds coincide. (The one additional experiment in Theorem **??** only makes a difference when $N$ is a power of 2. In those cases the adjacency information – present in an OME, but not assumed in Theorem **??** – cannot be established completely in the $\log_2(N)$ experiments.)

An algorithm that reduces by half the size of all cliques of undirected edges, for which no edge-orientation is known, clearly satisfies the conjectured bound. In general, the requirement is a little weaker: If all undirected cliques $C_{>h}$, that are larger than $h$ variables, where $h$ is the closest power of 2 below $|C_{\max}|$, are reduced to cliques of size $h$ in each experiment, then the conjectured bound is satisfied.

The conjecture currently remains without proof because it is not entirely clear whether it is possible to find intervention sets for any OME that break down all such cliques sufficiently fast. Since cliques may overlap, a proof of the conjecture must guarantee that intervention sets can always be found that resolve all of the overlapping cliques at once. In particular, consider a knowledge graph containing an undirected five-cycle as shown below. Here we have four overlapping maximal cliques of size 2. Again, one experiment should be sufficient to recover the causal structure, but in this case it is impossible to find an intervention set that resolves all four cliques simultaneously.

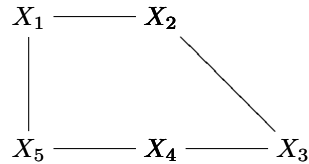

Any selection would result in one clique for which either both or no variable is contained in the intervention set. However, the five-cycle is not a counterexample to the conjecture since an undirected five-cycle (or any other cordless cycle greater than three) cannot occur in any OME – it would always imply a v-structure). Hence, the conjecture does not hold for arbitrary knowledge graphs, but is dependent on the assumption that the graph is an OME or derived from one.

For arbitrary knowledge graphs there is a general negative graph theoretic result due to Folkman (1970). It relates the problem of intervention set selection to coloring theorems:

**Theorem 2.2** (Folkman Clique Theorem – paraphrased). *For any clique-size $c \geq 3$, there is a graph $G$, whose largest clique has size $c$ and for which every edge two-coloring has a clique of size $c$ in one color.*

Considering only the undirected edges of an OME, let an edge be colored red in experiment $\mathcal{E}$ if it connects an intervened and a non-intervened variable, and blue if it connects variables that are both subject to an intervention or both passively observed. In some cases, only the red edges will be oriented as a result of the experiment. Folkman's theorem implies that for any integer $c \geq 3$, there is a graph $G$ whose largest clique has $c$ members. Any coloring, so in particular the coloring we defined, would result in a clique of size $c$ of blue or red edges only. Due to the way our coloring is defined, it is impossible for the clique to be among red edges, since three variables cannot be fully connected by red edges. Consequently, the clique is among blue edges, and since the intervened variables are separated from the non-intervened ones (by red edges), the clique must be either among the intervened variables only or among the unintervened variables only. That is, after the experiment, we may be left with a clique the same size as we started off with. Since there is no way to reduce cliques of un*known* edges by more than half, the conjectured bound does not hold for general knowledge graphs, not even for general adjacency graphs. However, our simulations give hope – as with the five-cycle – that the knowledge graphs that satisfy Folkman's theorem are not OMEs and not derivable from OMEs by sequences of experiments. If that fails, an argument is needed that such graphs are sufficiently rare so as not to be of practical worry.

## 3 ALGORITHM

The algorithm takes as input a knowledge graph over the set of variables $\mathbf{V}$. It associates with each vertex a boolean field that specifies whether the vertex is admissible to the intervention set or not. A vertex may not be admissible to an intervention set if it is part of a clique for which too many vertices are already part of the intervention set. The free parameter *maxInter* limits how many nodes may be simultaneously subject to an intervention in one experiment. As it stands, the algorithm is not optimal if $maxInter \leq N/2$, but see the discussion below.

**Algorithm 3.1** (OPTINTER: Intervention Set Selection). *Given a knowledge graph over a set of vertices $\mathbf{V}$, each vertex in $\mathbf{V}$ can be determined to be* admissible *or* inadmissible *and each vertex has a counter (of clique memberships). Let* maxInter *be the maximum size of the intervention set $\mathbf{I}$ for the next experiment.*

1. *Mark all vertices as* admissible *and set the counters for each vertex to 0.*

2. *Initialize the intervention set $\mathbf{I}$ to be the empty set.*

3. *Find all maximal cliques of vertices connected by* unknown *edges and order them $C_{\max}$ to $C_{\min}$ by the number of vertices they contain. (No need to resolve ties.)*

4. *Each clique can be either* resolved *or* unresolved. *Mark all maximal cliques as* unresolved.

5. *Compute $h = 2^{\lceil \log_2(|C_{\max}|)\rceil - 1}$ (the closest power of 2 with $2^n < |C_{\max}|$).*

6. *Let the* relevant *cliques $C_1, ..., C_k$ be the cliques with $|C_i| > h$.*

7. *Sort all relevant cliques in order of size, place among equal sized cliques the ones with the most* inadmissible *nodes first.*

8. *Let $C_{curr}$ be the first (largest) unresolved clique in the list of relevant cliques.*

9. *For each vertex $U \in C_{curr}$, set its counter to the number of* unresolved relevant *cliques $C_i$ it is part of.*

10. *While $(|\mathbf{I}| < maxInter) \&\& (|C_{curr} \cap \mathbf{I}| < |C_{curr}| - h)$, select vertex $V \in C_{curr}$ such that $V$ is* admissible *and has the highest count; select randomly among ties. Place it in $\mathbf{I}$.*

    (a) *For any relevant clique $C_i$, if $|C_i \cap \mathbf{I}| = |C_i| - h$, then mark $C_i$ as* resolved.

    (b) *For any relevant clique $C_i$, if $|C_i \cap \mathbf{I}| = h$, mark its vertices as* inadmissible.

11. *Return to 7 and start over until all relevant cliques are* resolved *or when no further relevant cliques can be resolved.*

12. *(Post Process: While possible with regard to the constraints (a) and (b) of step 10, add vertices to the intervention set to resolve additional maximal cliques.)*

13. *Return the intervention set.*

## 4 SIMULATION

Since we know that Conjecture ?? is not true for general knowledge graphs and we have reason to believe that it holds true for OMEs, we tested the conjecture in a simulation. Since the space of DAGs grows super-exponentially in the number $N$ of variables we have for even very small $N$ a space for which there is no obvious technique on how to generate a reasonable sample of graphs that could be called a representative test of the conjecture. In addition, if we just sampled randomly from the space of all possible DAGs over a set of variables, we would get a sample of graphs which – for the most part – would have very similar (and in fact quite small) maximum clique sizes. Using an MCMC technique to wait for a random DAG with a large clique and then test that, is not feasible given how rare graphs with large cliques are. Constructing graphs with large cliques deterministically runs the risk that our construction method may exclude those graphs that would be a counterexample to the conjecture.

So here is what we did: We considered DAGs with 12 vertices. This was the maximum $N$ for which we could sample and compute a large number of DAGs. We did not consider DAGs with fewer vertices since we assumed that these would occur as subgraphs of the 12-DAGs anyway. By sampling randomly from the space of all possible 12-DAGs we were easily able to generate a large enough number of DAGs whose OME contained maximal cliques with sizes between 1 and 4. To obtain samples of 12-DAGs with OMEs with larger cliques, we constructed 12-DAGs that were completely connected and then randomly deleted 2 edges, and then generated the OME from the remaining DAG. This yielded a decent number of OMEs with cliques between 3 and 10. 11-node cliques are impossible in 12-DAGs if two edges are deleted, but since there is only one DAG – modulo variable renaming – with a maximum clique containing 11 nodes, we confirmed the conjecture in this case by hand, similarly for the complete 12-DAG. (Consequently they are listed as only "1*" sample in the table below). Our sample is in no obvious sense representative, but it does contain randomly sampled graphs that imply OMEs with a large variety of different maximum clique sizes.

We used the OME generated from the independence constraints that a sampled 12-DAG implied as the knowledge graph at the outset of our sequence of experiments. We recorded the size of the largest clique in the OME, and then performed a sequence of experiments, in which for each experiment the intervention set was determined by OPTINTER. In each experiment we assumed that we had access to an oracle for the independence constraints of the manipulated distribution, and we updated our knowledge graph after each experiment given the new independence constraints. We recorded the number of experiments necessary to uniquely determine the DAG.

In the plot below we show the number of experiments required to uniquely determine the causal structure given the OME of 12-DAGs with various maximum clique sizes (x-axis). We plot the mean and the maximum number of experiments that were required, and show how they compare against the conjectured number of experiments. (Theoretically the minimum number of experiments is always 1 for maximal cliques greater or equal to two in an OME.)

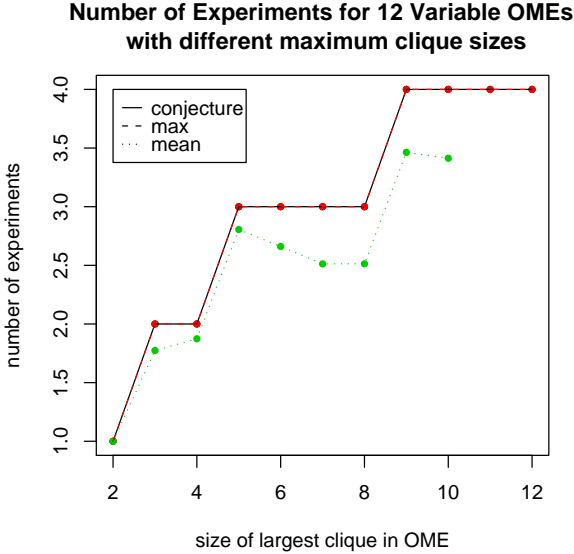

**Number of Experiments for 12 Variable OMEs with different maximum clique sizes**

The plot shows that the conjecture is not violated and that it in fact appears to be tight in the sense that the maximum number of experiments necessary to uniquely determine the causal graph does in fact coincide with the conjectured bound. Furthermore, we see that the mean number of experiments grows more slowly, despite the fact that we sampled very dense graphs (with 53 out of a possible 55 edges) to test maximal clique sizes of 5 and greater. OPTINTER selects variables for the intervention set randomly if there is no reason to prefer one over the other, and so the lower mean is due to "lucky" samples of intervention variables. Of course, the particular growth rate of the mean should not be taken as informative because of our sampling method. The mean is based on different sample sizes in each case. The following table shows how many OMEs each data point in the plot is based on:

| MaxClqSize | 1 | 2 | 3 | 4 | 5 | 6 |
|---|---|---|---|---|---|---|
| # of OMEs | N/A | 495 | 349 | 253 | 278 | 286 |
| MaxClqSize | 7 | 8 | 9 | 10 | 11 | 12 |
| # of OMEs | 242 | 152 | 123 | 46 | 1* | 1* |

## 5 DISCUSSION

So far we only considered experiments involving *multiple simultaneous* interventions on an OME. The difficulty in specifying a bound on the number of experiments given an OME, when only single (or only very few) interventions are permitted per experiment, is due to the interdependence of the orientation of two adjacent edges. For example, if we know from passive observation that the OME of the true causal graph is a chain of undirected edges,

$$X_1 \text{------} X_2 \text{------} \cdots \text{------} X_N$$

then we know that this chain cannot contain any v-structures, since they would have been discovered in the passive observation. But it could contain a common cause at any vertex (except the ends). Consequently, if only a single intervention can be performed per experiment, the most efficient strategy would be to intervene on the middle vertex, resulting in a sequence of $\log_2(N)$ experiments in the worst case to recover the causal structure. OPTINTER, without constraints on $maxInter$ would perform a single experiment with every other variable in the intervention set. With $maxInter = 1$, OPTINTER might well take $N/2$ experiments. Similar arguments apply to tree structures and particular planar networks made up of triangles or diamond shapes. Ultimately, if there are many dependencies between the directions of edges, then these can be exploited to improve the efficiency of discovery, but greedy procedures like OPTINTER are suboptimal. In general the number of experiments necessary and sufficient to discover the causal graph given an OME when only single interventions can be performed per experiment, is bounded by the $\sum_i(|C_i|-1)$, where the $C_i$ are *non-overlapping* maximal cliques. But we have no results on how tight this bound is, nor do we have a method, other than brute force, to compute the appropriate intervention sets. For similar reasons, the intervention sets computed by OPTINTER are not minimal with regard to the number of variables subject to an intervention in one experiment or over the entire sequence of experiments.

Computing the appropriate intervention set given a knowledge graph is closely related to the MAX-CUT problem, which is in general NP-complete. There are approximation algorithms, with the best offering a 0.878-approximation (Goemans & Williamson 1995). The approximation MAX-CUT algorithm is, of course, not designed with the specific aim to orient edges in cliques. Hence, an approximate MAX-CUT might not be sufficient to guarantee the conjectured bound (even if true). The OPTINTER algorithm is a greedy algorithm that selects the intervention set specifically in light of the above conjectured bound. But since OPTINTER is computationally expensive (due to the clique search), a MAX-CUT approximation algorithm may be a better choice for large graphs even if the guarantees supplied may be weaker than those of OPTINTER.

## 6 CONCLUSION

We have provided an algorithm for the computation of intervention sets based on the qualitative structural constraints on hypothesis spaces that can be represented by knowledge graphs. Knowledge graphs provide a concise representation of hypotheses that characterize the underdetermination of causal structures from passive observational data (observational Markov equivalence classes) and underdetermination resulting from sequences of experiments. We conjecture that the number of experiments sufficient and in the worst case necessary to discover the true causal structure in an OME is a function of the largest undirected clique in the OME, and we have shown why this conjecture cannot be generalized to all knowledge graphs. In simulations we have shown that our OPTINTER algorithm determines intervention sets that generate sequences of experiments that satisfy the conjecture. As a result, we have a search procedure that is adaptive and therefore much more efficient (in the number of experiments) than procedures that satisfy worst case bounds for the space of all DAGs. Consequently, we think this algorithm is much more relevant to actual scientific practice.

**Acknowledgements**

I am very grateful to Oleg Pikhurko for pointing me to Folkman's Theorem. This research was funded by a fellowship from the James S. McDonnell Foundation.